\documentclass[runningheads]{llncs}
\usepackage{mwe} 
\usepackage{lineno,hyperref}
\modulolinenumbers[5]
\usepackage{graphicx}
\usepackage{caption}
\usepackage{multirow}
\usepackage{amssymb}
\usepackage{amsmath}
\usepackage{amstext}
\usepackage{amsgen}
\usepackage{textcomp}
\usepackage{amsbsy}
\usepackage{amsopn}
\usepackage{mathrsfs}
\usepackage{tikz}
\usepackage{balance}
\usepackage{hyperref}
\usepackage{booktabs}
\usepackage{verbatim}
\usepackage{array} 
\usepackage{picture}
\usepackage{backrefbis}
\usepackage{xfrac}
\usepackage{xspace}
\usepackage{review} 
\usepackage{soul}
\makeatletter 
\newcommand\mynobreakpar{\par\nobreak\@afterheading} 
\makeatother

\newcommand{\beq}{\begin{equation}}
\newcommand{\eeq}{\end{equation}}

\newcommand{\omitme}[1]{}

\newcommand{\confusedHL}[1]{}
\newcommand{\confusedJM}[1]{}

\usepackage{ulem}

\begin{document}
%


\title{Semi-supervised few-shot learning for medical image segmentation}


%
%
\author{Abdur R Fayjie\inst{1,2} \and
Reza Azad\inst{1} \and
Marco Pedersoli\inst{1} \and
Claude Kauffman\inst{2} \and
Ismail Ben Ayed\inst{1} \and
Jose Dolz\inst{1}}
\authorrunning{Fayjie et al.}
%
\institute{ETS Montreal, Montreal, Canada \and
CRCHUM Montreal, Montreal, Canada
 \\
\email{abdur-razzaq.fayjie.1@ens.etsmtl.ca}}
\maketitle              
\begin{abstract}
Recent years have witnessed the great progress of deep neural networks on semantic segmentation, particularly in medical imaging. Nevertheless, training high-performing models require large amounts of pixel-level ground truth masks, which can be prohibitive to obtain in the medical domain. Furthermore, training such models in a low-data regime highly increases the risk of overfitting. Recent attempts to alleviate the need for large annotated datasets have developed training strategies under the few-shot learning paradigm, which addresses this shortcoming by learning a novel class from only a few labeled examples. In this context, a segmentation model is trained on episodes, which represent different segmentation problems, each of them trained with a very small labeled dataset. In this work, we propose a novel few-shot learning framework for semantic segmentation, where unlabeled images are also made available at each episode. To handle this new learning paradigm, we propose to include surrogate tasks that can leverage very powerful supervisory signals --derived from the data itself-- for semantic feature learning. We show that including unlabeled surrogate tasks in the episodic training leads to more powerful feature representations, which ultimately results in better generability to unseen tasks. We demonstrate the efficiency of our method in the task of skin lesion segmentation in two publicly available datasets. Furthermore, our approach is general and model-agnostic, which can be combined with different deep architectures.  

\end{abstract}

Semantic segmentation is of vital importance in medical imaging, as it can assist in the treatment, diagnosis and follow-up of many diseases. Despite the automation of this task has been widely studied in the last decades, the recent advances in deep learning are driving progress on this problem. Particularly, Convolutional Neural Networks (CNN) have achieved state-of-the-art performance in a breadth of medical image segmentation problems, such as brain tissue \cite{chen2018voxresnet,dolz20183d}, heart structures \cite{bernard2018deep}, and abdominal organs \cite{fechter2017esophagus}. Nevertheless, a main limitation of these high-performing models is the strong need of large labeled datasets for training, which hampers their scalability to novel or rare categories.

To alleviate this issue, few-shot learning \cite{ravi2016optimization} has recently emerged as an efficient alternative to traditional fully supervised learning strategies. In this context, the CNN is trained to learn novel categories with only a few labeled images, which are typically referred to as \textit{support} images. Then, the knowledge derived from the \textit{support} images is employed to guide the segmentation of images containing the novel classes, known as \textit{queries}. To reproduce the scenario found during testing, the network is trained on the labeled examples following the episodic training paradigm \cite{vinyals2016matching}. This is, at each step we sample $k$ labeled examples from $n$ novel categories to form an episode, which is used to train the segmentation network. By doing this, the network is trained to extract useful information for all the different episodes but prevents the specialization of a particular novel task. Recent techniques to improve the generability on this scenario integrate a mask average pooling strategy, that masks out irrelevant features based on the \textit{support} masks \cite{nguyen2019feature,siam2019amp,zhang2018sg}. Wang et al. \cite{wang2019panet} further improves generability to new classes with an additional novel prototype alignment regularization between support and query images. In other recent works \cite{hu2019attention,zhang2019canet}, deep attention has been exploited to learn attention weights between support and query images for further label propagation. Nevertheless, these methods do not leverage unlabeled data during training, which may help in low-labeled data scenarios.

Furthermore, despite the satisfactory results achieved by this new learning paradigm on the segmentation of natural images \cite{reza2020dog,nguyen2019feature,siam2019amp,wang2019panet}, its use in medical images remains scarce. Few-shot segmentation on medical images was first introduced by the work in \cite{mondal2018few}. Authors proposed to leverage adversarial learning to segment brain images based on 1 or 2 labeled brain images, inspired by the success of prior semi-supervised approaches \cite{souly2017semi}. Roy et al. \cite{roy2020squeeze} presented a different approach, where the architecture was composed of a conditioner and a segmenter arm. To strength the information exchange between both arms, they integrated \textit{squeeze} $\&$ \textit{excite} modules \cite{hu2018squeeze}, which facilitated the gradient flow. More recently, one-shot medical image segmentation was addressed by synthesizing realistic training examples \cite{zhao2019data}, a more elegant way of performing data augmentation. Nevertheless, these methods present some limitations. First, these approaches are based on the assumption that each shot is a whole 3D image, which contains many 2D slices. And second, they integrate large architectures (e.g., encoder, decoder and discriminator in \cite{mondal2018few} or conditioner and segmenter arms in \cite{roy2020squeeze}), which incur in complex and potentially unstable models. 

Inspired by this, our work investigates the role of unsupervised data, via surrogate tasks, in the task of segmenting medical images in a few-shot learning scenario. Particularly, we take advantage of the success of few-shot segmentation works in natural images, which is based on the episodic training paradigm. To the best of our knowledge, this is the first attempt to tackle the few-shot segmentation task in medical imaging from an episodic perspective. To further improve the performance of our model, we leverage unlabeled data as a supervisory signal of surrogate tasks. Integration of unlabeled images into auxiliary tasks, has already shown to improve the generalization capabilities of deep models in several visual recognition tasks in other domains \cite{doersch2015unsupervised,gidaris2019boosting}. To evaluate the effectiveness of the proposed approach we resort to the task of skin cancer segmentation in two publicly available medical datasets. Results demonstrate that our approach brings an improvement of 6-7\% over the baseline without incurring in extra-costs due to data annotation.



\begin{figure}[t]
\centering
\vspace{-5mm}
\includegraphics[width=0.95 \textwidth, trim=0.2cm 4.5cm 0.2cm 0cm]{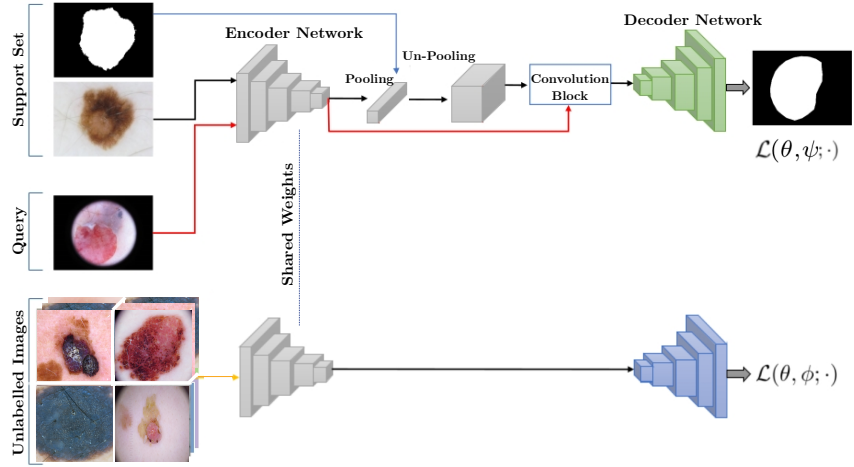}
\vspace{-4mm}
\caption{Proposed few-shot semantic segmentation method for medical images using auxiliary tasks to leverage abundant unlabeled medical imaging data.} 
\label{fig:method}
\end{figure}


\section{Methodology}

\subsection{Few shot semantic segmentation} 

In standard few-shot semantic segmentation, we typically have three datasets: a training set $D_{train}=\{(X_i^t,Y_i^t)\}_{i=1}^{N_{train}}$, a support set $D_{support}=\{(X_i^s,Y_i^s)\}_{i=1}^{N_{support}}$, and a test set $D_{test}=\{(X_i^q)\}_{i=1}^{N_{test}}$. In this setting, we denote an input image as $X_i \in \mathbb{R}^{H\times W \times Ch}$, where $H$, $W$ and $Ch$ represent the height, width and the number of channels, respectively, and $Y_i \in \{0,1\}^{H\times W}$ its corresponding pixel-level mask. In total, each dataset contains $N$ images, defined by $N_{train}$, $N_{support}$, and $N_{test}$, with $C$ different classes. While classes are shared among support and test sets, they are disjoint with the training set, i.e.,  $\{C_{train}\} \cap \{C_{support}\}=\emptyset$.

Few-shot learning aims at training a neural network $f_{(\theta,\psi)}(\cdot)$ on the training set to have the ability to segment a novel class $c \notin C_{train}$ on the test set based on $k$ references from $D_{support}$. Note that we employ $\theta$ and $\psi$ to refer to the learnable parameters of the encoder and decoder, respectively. To reproduce this procedure, training on the base dataset $D_{train}$ follows the episodic learning paradigm proposed in \cite{vinyals2016matching}, where each episode instantiates a $c$-way $k$-shot learning task. Specifically, each episode is generated by sampling two elements. First, we build a support training set for each class $c$, denoted as $D_{train}^{\mathcal{S}}=\{(X_s^t,Y_s^t(c))\}_{s=1}^{k} \subset D_{train}$, where $Y_s^t(c)$ is the binary mask for the class $c$ corresponding to the support image $X_s^t$. And second, a query set $D_{train}^{\mathcal{Q}}=\{X_q^t,Y_q^t(c)\} \subset D_{train}$, where $X_q^t$ is the query image and $Y_q^t(c)$ its corresponding binary mask for the class $c$. To estimate the segmentation mask of a given class $c$ in the query image the model gets the support training set and the query image, which can be expressed as $\hat Y_q^t(c)=f_{(\theta,\psi)}(D_{train}^{\mathcal{S}},X_q^t)$.

Concretely, we first employ a CNN to encode the support and query images into the feature space, resulting in $f_s \in \mathbb{R}^{W' \times H' \times M}$ and $f_q \in \mathbb{R}^{W' \times H' \times M}$, respectively. The variables $W'$, $H'$ and $M$ denote the width, height and feature dimensionality on the feature space, respectively. To obtain the class prototypes, we apply mask average pooling on the feature representation over the known foreground regions of the support mask ${Y}^t_s(c)$. This can be formulated as:

\begin{equation}
p_s = \frac{1}{| \tilde{Y}^t_s(c)|} \sum_{i=1}^{W' \times H'}f_s\tilde{Y}^t_s(c)
\end{equation}

where $\tilde{Y}^t_s(c) \in \{ 0,1\}^{H' \times W'}$ denotes the down-sampled version of the support mask $Y^t_s(c)$ and $|\tilde{Y}^t_s(c)|=\sum_i\tilde{Y}^t_{s,i}(c)$ is the number of foreground locations in $\tilde{Y}^t_s(c)$. Note that $p_s$ is a unidimensional vector with $M$ elements, $p_s \in \mathbb{R}^{1\times 1\times M}$. Then, we unpool each prototype to the same spatial resolution as the query features $f_q$ and convolve the upsampled prototypes with $f_q$ (See Fig. \ref{fig:method} for the whole pipeline). The model parameters $(\theta,\psi)$ are then optimized by employing an objective function between $Y_q^t(c)$ and $\hat Y_q^t(c)$. In the few-shot segmentation literature, this function is typically the standard cross-entropy. Nevertheless, any other loss function could be used. Last, during testing, the model $f_{(\theta,\psi)}(\cdot)$ is evaluated on the test set $D_{test}$ given $k$ images from the support set $D_{support}$.

\subsection{Surrogate task}

A major challenge that we can encounter in few-shot learning is how to force the feature extractor to learn image features that can be readily employed on novel classes with just a handful of labeled samples. To address this issue, we propose to integrate an auxiliary loss during training, which compensates for the lack of annotated data on the novel categories. We can define formally this loss as $\mathcal{L}(\theta,\phi;\cdot)$, where $\phi$ denotes the parameters of the network related to the surrogate task. 

\paragraph{\textbf{Integrating unlabeled data.}} Several techniques have been proposed for self-supervised training, including predicting rotation \cite{gidaris2018unsupervised}, solving jigsaw puzzles \cite{noroozi2016unsupervised} or filling removed parts of an image \cite{zhang2017split}. Among these strategies, we consider the task of image denoising, since some other techniques are not suited to our application. For example, if an image containing a car is rotated 180 degrees, it will be easy to predict its rotation, as the wheels would typically not be above the car roof. In contrast, an image of a skin tumor can be rotated in many directions, resulting in a large range of feasible rotations. To achieve this, we employ an additional dataset, $D_{train}^\mathcal{U}=\{(X_i^\mathcal{U})\}_{i=1}^{N_{\mathcal{U}}}$ , which contains $N_{\mathcal{U}}$ unlabeled images. We add random noise to these images, resulting in a larger dataset, where each image $X_i^\mathcal{U}$ generates multiple corrupted images $Y_i^\mathcal{U}$. The goal is that the encoder-decoder architecture learns a mapping between noised images and their original counterparts. To this end, we use the cross-entropy loss between the original and denoised images, which can be defined as: 
\begin{equation}
\mathcal{L}_{sur}(\theta,\phi;Y^\mathcal{U}) = - \frac{1}{N_\mathcal{U}}\sum_{i}^{N_\mathcal{U}} \sum_{j}^{H\times W} X_{i,j}^\mathcal{U} log \hat Y_{i,j}^\mathcal{U}
\label{eq:denoiseLoss}
\end{equation}
where $\hat Y_{i}^\mathcal{U}=f_{(\theta,\psi)}(Y_{i}^\mathcal{U})$ is the denoised image. We employed the CE instead of L$_2$ distance as the reconstruction loss as we empirically observed that it provided better results. This is in line with recent literature in Variational Auto-Encoders \cite{basu2019early}, which also employ CE as the reconstruction loss.

\subsection{Joint objective}

The final objective optimized during training is thus composed by the two terms included in the previous sections. The first term $\mathcal{L}_{few}(\theta,\psi;\cdot)$ is a function of the parameters $\theta$ and $\psi$ of the encoder and decoder specialized on the few-shot segmentation task. 
The second term, $\mathcal{L}_{sur}(\theta,\phi;\cdot)$, depends on the encoder parameters $\theta$ and on the parameters $\phi$ of a network only dedicated to the surrogate task. Thus, the training process reduces to minimize the following function:

\begin{equation}
    \min_{\theta,\psi,\phi} \mathcal{L}_{few}(\theta,\psi;D_{train}) + \lambda \mathcal{L}_{sur}(\theta,\phi;Y^\mathcal{U})
    \label{total_loss}
\end{equation}

where $\lambda$ is employed to weight the importance of the surrogate task.

\section{Experiments}
We conduct a series of experiments to evaluate the proposed model for few-shot segmentation. We present below the datasets and experimental settings. 

\vspace{-5mm}

\subsection{Dataset}
We employ the FSS-1000 dataset as the base training set ($D_{train}$) in our experiments. To evaluate our method, we employ two publicly available medical datasets, i.e., ISIC and PH$^2$, which are described below. For the surrogate tasks, we only employ images from both FSS-1000 and ISIC datasets.

\noindent
\textbf{\textit{FSS-1000 Class Dataset: }}
FSS-1000 class dataset \cite{wei2019fss} is a large-scale dataset specially designed for few-shot segmentation. It consists on 1000 classes, where each class contains 10 images with their corresponding pixel level ground truth annotations. The official training split (760 classes) is used as the base dataset for training, while the testing set (240 classes) is exploited on the surrogate tasks.

\noindent
\textbf{\textit{ISIC dataset: }}
The ISIC 2018 dataset \cite{codella2019skin,tschandl2019ham10000} is provided by the International 2018 Skin Imaging Collaboration Grand Challenge. The dataset contains 2594 dermoscopic images in RGB with their respective masks. An independent set of 1000 images is kept for evaluation purposes. The remaining images are employed for the auxiliary tasks, except the images used as $k$-shots in the support set. We resize all the images to 224 $\times$ 224 pixels to match the resolution of the images in the FSS-1000 dataset.

\noindent
\textbf{\textit{PH$^2$ dataset: }} The PH$^{2}$ dataset \cite{mendoncca2013ph} contains a total of 200 RGB dermoscopic images of melanocytic lesions obtained at the Dermatology Service of Hospital, Pedro Hispano (Matosinhos, Portugal). All these images are used only during testing. 
Similarly to ISIC, we resized the images to 224 $\times$ 224 pixels.

\subsection{Experimental Set-up}

\textbf{\textit{Network Details: }} Our encoder consists of four encoding blocks from VGG \cite{simonyan2014very} pre-trained on ImageNet. We removed the max-pooling layers of the last two blocks and used atrous convolutions with dilation rate of 2 to enlarge the receptive field. 
Size of input images is equal to $224\times224$ and their encoded representations are down-scaled to 1/4 of the original input resolution. Our decoder network consists of upsampling, convolution, batch normalization and activation layers. To generate the final segmentation mask, we apply two decoding blocks.

\noindent
\textbf{\textit{Evaluation Protocol: }} In our work, we evaluate the performance of our model based on the DSC score, widely used in medical imaging to evaluate the segmentation performance. Given two segmentation masks A and B, the DSC can be defined as $DSC  =  \frac{2\left|A \cap B \right|}{|A|+|B|}$.



\noindent
\textbf{\textit{Implementation Details: }} 
The code is written in Keras with Tensorflow as backend. The tests are carried out in a server equipped with a Nvidia Titan X GPU. Both main and surrogate tasks are trained end-to-end using Adam optimization with learning rate 10$^{-4}$. 
In whole setting we trained the model for 30K iterations and evaluated on 500 episodes. The value of $\lambda$ in eq. \ref{total_loss} is set 1.

\subsection{Results}
\paragraph{\textbf{Quantivative results.}}Table \ref{tab:one_shot} reports the segmentation results on ISIC and PH$^2$ datasets in the 1-shot scenario. First, to show the effectiveness of few-shot learning, we start by comparing this setting with standard batch-wise training, referred to as \textit{Regular}. In this case, the model is trained on FSS-1000 dataset and directly tested on both ISIC and PH$^2$ sets. We can observe that across the two datasets, resorting to the episodic training paradigm results in an improvement of nearly 6\%. Looking at the results obtained by our model, these indicate that leveraging unsupervised data via a surrogate task consistently improves the performance on both datasets. Particularly, the proposed approach outperforms the few-shot baseline by a margin between 6-7\%. Compared to the upperbound, i.e., model trained and tested on the same dataset, the proposed models obtain promising results, with around 15\% of difference on the PH$^2$ dataset, but only 1 target image segmented. Furthermore, we can also observe that increasing the number of additional samples on the episodes typically leads to better results. This suggests that the model efficiently extracts semantic information from the unlabeled surrogate task to enhance the performance of the downstream task.

\begin{table}[h!]
\setlength{\tabcolsep}{6pt}
\centering
\begin{tabular}{l|c|c|c}
\toprule
\textbf{Model}     & \textbf{\begin{tabular}[c]{@{}c@{}}Additional\\ Samples\end{tabular}} & \textbf{ISIC} & \textbf{PH$^2$} \\
\hline
Regular (\textit{Lower-bound})     &  ---        &  48.32   &  64.72          \\
\hline
Few-shot     &  ---        &  54.07   &   68.13             \\
\hline
\multirow{3}{*}{\begin{tabular}[c]{@{}l@{}}Few-shot + unlabeled\\ \textit{(Denoising}) \textbf{\textit{(Ours)}}\end{tabular}}       &   5         &  61.38   &    74.12      \\
    &  10         &  \textbf{61.40}   &    74.67             \\
     &  20         &  60.79   &    \textbf{74.77}          \\
     \hline
Regular (\textit{Upper-bound})    &  ---        &   86.65    &     89.94     \\
\bottomrule
\end{tabular}
\caption{Quantitative results of the evaluated settings on ISIC and PH$^2$ datasets. Best results (different from the upperbound) highlighted in bold.}  
    \label{tab:one_shot}
\end{table}



Results for the 5-shot scenario are reported in Table \ref{tab:five_shot}. In this setting, we only evaluate the model when 10 additional images are integrated in the episodes. Similarly to the 1-shot case, we observe an increase in performance with respect to the few-shot model trained without the surrogate task. 
Concretely, the proposed model obtains an improvement of 3-4\%, whereas the gain was nearly to 6-7\% in the case of 1-shot learning. 

\begin{table}[]
\setlength{\tabcolsep}{6pt}
\centering
\begin{tabular}{l|c|c}
\toprule
\textbf{Model}                       & \textbf{ISIC} & \textbf{PH$^2$} \\
\midrule
Few-shot                             &     59.63   &   71.15    \\
Few-shot + Unlabeled \textit{(Denoising)} \textit{\textbf{(Ours)}}                &    \textbf{62.40}    &   \textbf{75.54}    \\
\bottomrule
\end{tabular}
\caption{Quantitative results of 5-shot settings on ISIC and PH$^2$ datasets. Best results highlighted in bold.}  
    \label{tab:five_shot}
\end{table}


These results suggest that adding a self-supervised surrogate task, i.e., denoising, improves the few-shot segmentation performance. Results also indicate that the performance improvement is more significant in the context of very few labeled samples, such as 1-shot \textit{vs.} 5-shot.



\begin{figure}[!h]
\centering
\vspace{-5mm}
\includegraphics[width=0.9 \linewidth]{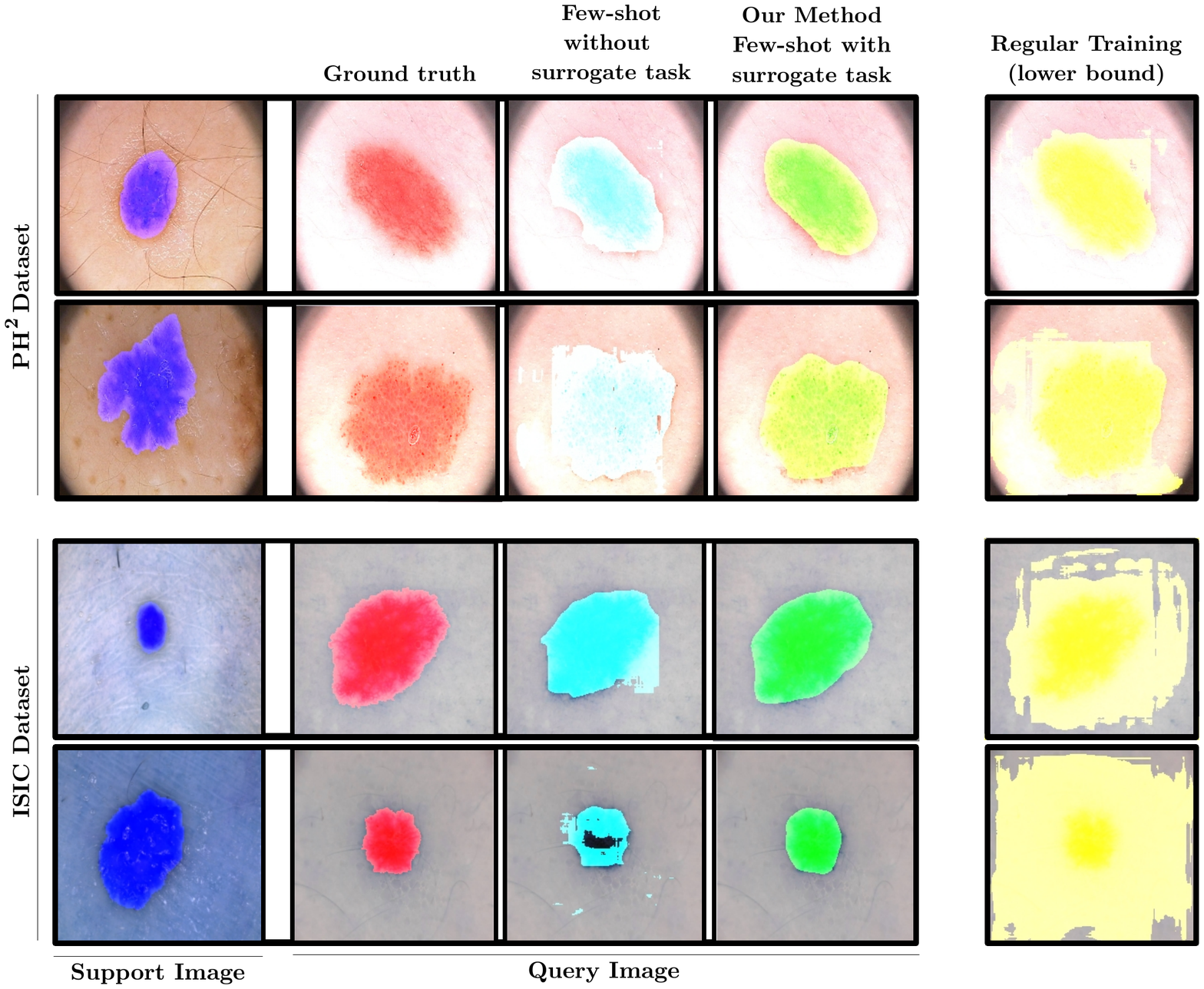}
\vspace{-4mm}
\caption{Visual segmentations of the analyzed methods in the scenario of one-shot segmentation, on \textit{PH$^2$} and \textit{ISIC} datasets. } 
\label{fig:miccai_ph}
\end{figure}

\paragraph{\textbf{Qualitative results.}} Visual segmentations on both datasets are depicted in Fig. \ref{fig:miccai_ph}. We can first observe that, by training the network  in a batch-wise manner (\textit{right column}) the segmentation results are not satisfactory, largely oversegmenting the target. If episodic training is used instead, the network shows a stronger capability to learn more general features. This is reflected in the better segmentation results provided by the few-shot baseline (\textit{third column}). Last, leveraging unlabeled data through a surrogate task further improves the representation power of the model, resulting in richer and more generic features (\textit{fourth column}). Learning better features ultimately results in more reliable segmentations, as demonstrated in these visual examples. Specifically, the proposed model improves the segmentation by increasing the number of true positives (e.g., whole in the target of last row), while reducing the amount of false positives (e.g., isolated pixels on last row, and over-segmentations of other examples).

\vspace{-5mm}
\section{Conclusion}
Inspired by the success of few-shot learning on visual recognition tasks we have proposed the first attempt to integrate episodic training in few-shot semantic segmentation on medical images. Furthermore, motivated by the close connection between few-shot and self-supervised learning, we have investigated the use of surrogate tasks to further improve current few-shot segmentation approaches. By solving a non-trivial proxy task that can be supervised trivially, such as denoising images with noise, the encoder network is encouraged to learn rich and generic image features which can be transferable to other ensuing tasks such as image segmentation. These enriched features lead to more powerful representations, which ultimately results in better generability to unseen tasks. Our experiments on two public skin cancer segmentation datasets revealed that exploiting unlabeled images through self-supervised tasks results in significant improvements on the few-shot segmentation performance.

%
%
%
%
%
\bibliographystyle{splncs04}
\bibliography{miccai-bib}
%


\end{document}